\documentclass[lettersize,journal]{IEEEtran}
\usepackage{amsmath,amsfonts}
\usepackage{algorithmic}
\usepackage{algorithm}
\usepackage{array}
\usepackage{textcomp}
\usepackage{stfloats}
\usepackage{url}
\usepackage{verbatim}
\usepackage{graphicx}
\usepackage{xcolor}
\usepackage{cite}
\usepackage{multirow}
\usepackage{booktabs}
\usepackage{subcaption}
\usepackage{float}
\newcommand{\RNum}[1]{\uppercase\expandafter{\romannumeral #1\relax}}
\begin{document}

\title{Real-Time Aerial Fire Detection on Resource-Constrained Devices Using Knowledge Distillation}

\author{Sabina Jangirova, Branislava Jankovic, Waseem Ullah, Latif U. Khan, Mohsen Guizani}
        % <-this % stops a space



\maketitle

\begin{abstract}

Wildfire catastrophes cause significant environmental degradation, human losses, and financial damage. To mitigate these severe impacts, early fire detection and warning systems are crucial. Current systems rely primarily on fixed CCTV cameras with a limited field of view, restricting their effectiveness in large outdoor environments. The fusion of intelligent fire detection with remote sensing improves coverage and mobility, enabling monitoring in remote and challenging areas. Existing approaches predominantly utilize convolutional neural networks and vision transformer models. While these architectures provide high accuracy in fire detection, their computational complexity limits real-time performance on edge devices such as UAVs. In our work, we present a lightweight fire detection model based on MobileViT-S, compressed through the distillation of knowledge from a stronger teacher model. The ablation study highlights the impact of a teacher model and the chosen distillation technique on the model's performance improvement. We generate activation map visualizations using Grad-CAM to confirm the model's ability to focus on relevant fire regions. The high accuracy and efficiency of the proposed model make it well-suited for deployment on satellites, UAVs, and IoT devices for effective fire detection. Experiments on common fire benchmarks demonstrate that our model suppresses the state-of-the-art model by 0.44\%, 2.00\% while maintaining a compact model size. Our model delivers the highest processing speed among existing works, achieving real-time performance on resource-constrained devices.
\end{abstract}


\section{Introduction}
Over many years, fire remains one of the most significant natural disasters, dangerous in its destructive character and speed of spreading. In 2023, the northern parts of Kazakhstan experienced massive wildfires, which burned more than 60,000 hectares of forest and killed 15 people \cite{Eurasianet}. Together with the loss of lives and environmental damage, fires lead to financial harm. These consequences can be reduced by early detection and correct classification of the ignited fire, ensuring a reactive response. Nowadays, intelligent fire detection systems are mainly deployed on CCTV cameras, which have a fixed line of vision and position. Therefore, there is a high risk of missing the start of the fire if it is located in the "blind" spot not covered by the cameras. Thus, there is still a need for more reliable solutions capable of classifying different types of fire under varied conditions and complex environments. By using unmanned aerial vehicles, we can monitor much larger, distant territories and also come closer to suspicious objects if it is hard to classify them as fire or non-fire. The constraint of such devices is that they have limited storage and computational capabilities. Therefore, they won't be able to utilize computer vision models with large, complex architectures. Motivated by the described challenges, our research takes advantage of state-of-the-art (SOTA) vision transformers and knowledge distillation (KD) techniques to enhance the accuracy and efficiency of fire detection systems on remote sensing devices, contributing to more effective fire prevention and management.

In the early stages, fire detection was performed using scalar sensor-based methods, such as the installation of smoke, particle, temperature, and flame detection sensors \cite{yar2021towards}. These methods are cheap and easy to install, but scalar sensors can monitor only indoor environments and thus have limited usage scenarios. Vision sensor-based methods work with video and image data, presenting a broad region coverage, reduced human intervention requirements, rapid response times, environmental resilience, and additional information on fire characteristics (e.g. the size of the affected area). Mostly, conventional machine learning (CML) and deep learning (DL) models are used for these methods \cite{yar2024efficient}. CML methods commonly employ features such as motion, color, shape, and texture \cite{harkat2023fire, xu2024detecting}, and the performance of such models is correlated with the quality of the features. Moreover, they failed to generalize to cases with poor weather conditions and complex, unseen scenarios.

Alternatively, DL methods proved to be effective in extracting characteristics, especially for the fire detection task \cite{muhammad2019efficient, khan2023performance, sathishkumar2023forest, wang2023domain, jin2024self, rui2023rgb}. A more complex architecture of such models allows for capturing intricate patterns and dependencies from the images. DL methods have demonstrated a prominent ability to enhance classification performance in adverse weather conditions and complex scenes, further validating their use despite the increased computational demands. 

Although DL models reduce false alarm rates compared to CML models, they require heavy computations and have limited capabilities to distinguish between fire and fire-like objects \cite{chino2015bowfire}. Many researchers developed solutions to overcome these limitations (\cite{yar2024efficient, yar2022optimized, park2022advanced, khan2023enhancing, sharma2017deep, li2020efficient, muhammad2018convolutional, daoud2023fireclassnet, dilshad2024toward, yar2024modified}), yet there's usually a trade-off between the model's processing speed and the accuracy of predictions. In practical scenarios, quick detection and correct response to the forming fire is necessary to prevent significant losses. This limitation drives the demand for solutions that balance accuracy with efficiency, enabling their deployment in real-time applications.

To address these challenges, we propose a novel approach utilizing MobileViT-S \cite{mehta2021mobilevit} as the best backbone model optimized with KD techniques. Our approach combines the precision of large-scale teacher models with the compactness and efficiency of student models, allowing for robust and scalable fire detection systems.
The main contributions of this work include:
\begin{itemize}
    \item We develop a model for fire detection employing the best backbone and implementing a KD approach, transferring crucial insights from a larger teacher model to a lightweight student model.
    \item We conduct an extensive ablation study to evaluate the effectiveness of the proposed teacher model, student model, and KD technique for performance. 
    \item We evaluated the performance of our model on three fire classification benchmarks. Our model achieves the same results and even exceeds the accuracy of the existing methods while being significantly more compact.
    \item We demonstrate our model's ability to focus on relevant areas within the images by using the Grad-CAM tool. Our proposed method provides meaningful information on the decision-making process of our proposed model, increasing its explainability. 
\end{itemize}

Section \RNum{2} describes the related work in the domain. Section \RNum{3} contains the framework proposed in this project, while Section \RNum{4} discusses the experiments and their results and presents the ablation study on the effect of different teacher models and KD techniques. Finally, Section \RNum{5} presents the conclusion of the work done in this research, possible implications, and future directions.

\section{Related Work}

Early CML methods for fire detection relied on color analysis and image processing techniques to extract fire and smoke features \cite{chen2004early, marbach2006image, rafiee2011fire, celik2009fire, borges2010probabilistic}, while later approaches integrated motion features, such as optical flow and spatiotemporal analysis \cite{foggia2015real, chen2010multi, ha2012vision}. Chen \textit{et al.} \cite{chen2004early} proposed a method that used color segmentation in the RGB color space to isolate fire-like regions in images, while Marbach \textit{et al.} \cite{marbach2006image} explored dynamic color modeling to adapt to varying fire hues. Celik and Demirel \cite{celik2009fire} introduced a statistical model for fire pixel detection based on brightness and color properties. Borges and Izquierdo \cite{borges2010probabilistic} took a probabilistic approach, combining color and temporal information for fire classification. However, these methods suffered from high false alarm rates due to the diverse characteristics of fire. To improve robustness, later methods incorporated motion features. Foggia \textit{et al.} \cite{foggia2015real} employed optical flow to capture the dynamic nature of fire, while Chen \textit{et al.} \cite{chen2010multi} used spatiotemporal analysis to distinguish between fire and non-fire motion patterns. Ha \textit{et al.} \cite{ha2012vision} combined motion and texture features to enhance detection reliability. Recently, Xu \textit{et al.} combined a Modified Pixel Swapping Algorithm with mixed-pixel unmixing and threshold-weighted fusion to detect forest fires, which improved accuracy and reduced false alarms \cite{xu2024detecting}. Although these methods reduced false alarms to some extent, they struggled in scenarios with camera movements or other moving objects that could mimic fire behavior.

Deep learning (DL) methods, particularly convolutional neural network (CNN) models, have shown improved performance in fire detection \cite{frizzi2016convolutional, lee2017deep, sharma2017deep}. Lightweight CNNs like those proposed by Muhammad \textit{et al.} \cite{muhammad2019efficient} and Daoud \textit{et al.} \cite{daoud2023fireclassnet} addressed computational constraints, but challenges in detecting small, distant fires in adverse conditions remain. Some researchers employed CNN-based models with attention mechanisms. Li \textit{et al.} proposed a fire detection approach with multiscale feature extraction, deep supervision, and channel attention mechanism \cite{li2020efficient}. Wang \textit{et al.} proposed a Dynamic Equilibrium Network to detect fire based on the data from different types of sensors \cite{wang2023domain}. In \cite{yar2022optimized}, the authors integrated the spatial attention (SA) and channel attention (CA) modules into the Inceptionv3 architecture and improved the performance of the backbone model. Similarly, \cite{khan2023enhancing} introduced the SA and CA modules to the ConvNeXtTiny architecture. Yar \textit{et al.} \cite{yar2024efficient} proposed a modified MobileNetV3 architecture with an added Modified Soft Attention Mechanism (MSAM) and 3D convolutional operations. Dilshad \textit{et al.} \cite{dilshad2024toward} developed an optimized fire attention network (OFAN) that consisted of a MobileNetV3Small as a backbone model, CA and SA mechanisms to capture global dependencies. Rui \textit{et al.} developed a multi-modal RGB-T wildfire segmentation framework that learns both modality-specific and shared features via parallel encoders and a shared decoder \cite{rui2023rgb}. Alternatively, Yar \textit{et al.} \cite{yar2024modified} proposed a ViT-inspired model with a shifted patch tokenization (SPT) module for spatial details, a locality self-attention (LSA) module to optimize the softmax temperature, and dense layers instead of the multi-head to reduce the complexity of the model. However, these methods still need more robustness and capability to capture small fire regions in complex scenes, like fog or hazy weather.

The challenge of detecting small, distant fire sources in poor weather persists in the current research. While CNNs with attention modules enhance feature representation through channel-wise attention, they primarily focus on local spatial information and channel dependencies. To address this limitation, we utilize the architecture that combines the efficiency of CNNs with the global modeling capabilities of Vision Transformers \cite{mehta2021mobilevit}.

\section{Proposed Framework}

\begin{figure*}
    \centering
    \includegraphics[width=\linewidth]{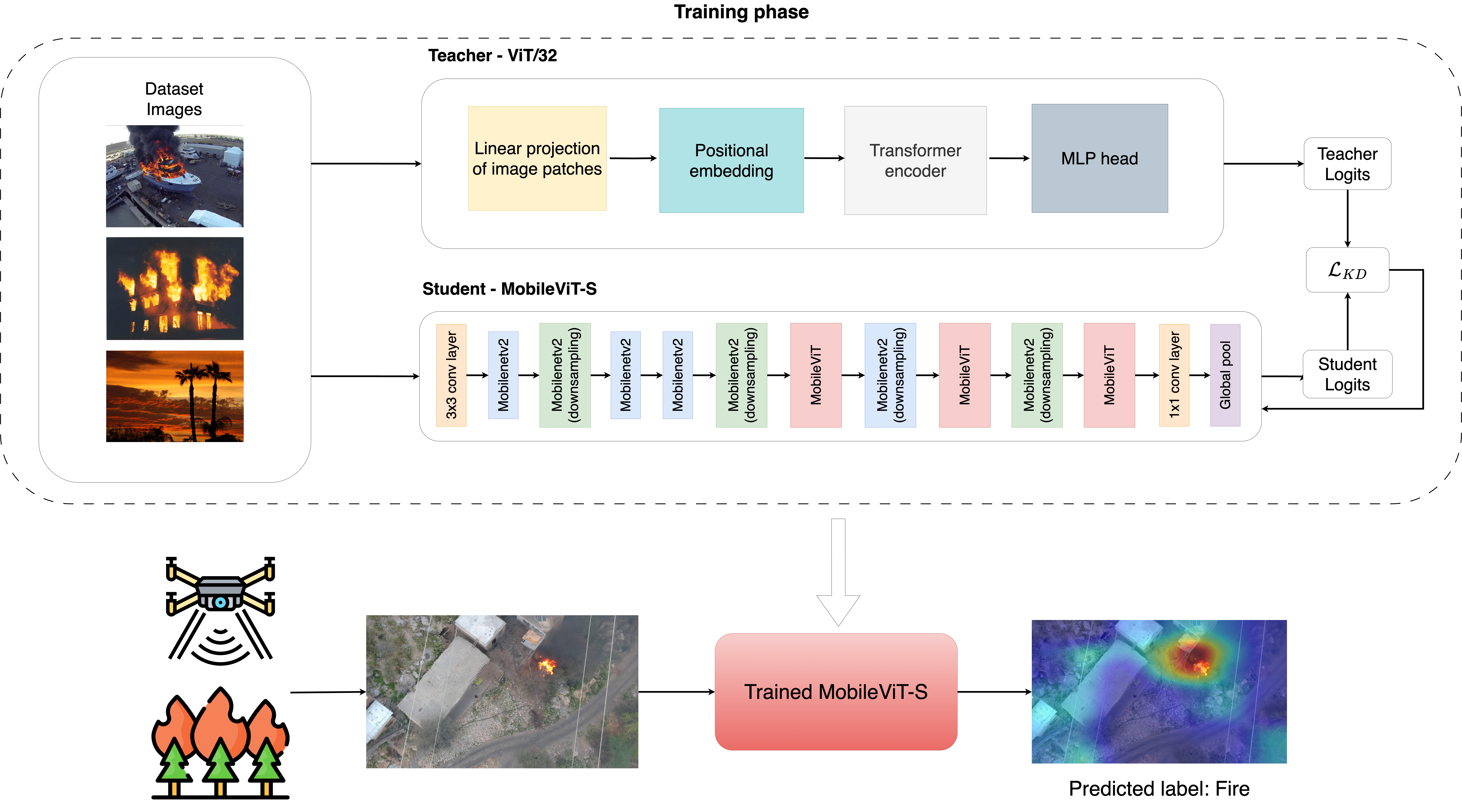}
    % \caption{Proposed framework. The depicted training phase involves distilling knowledge from ViT/32 to MobileViT-S, where $\mathcal{L}_{KD}$ is KD loss. The trained MobileViT-S is deployed on resource-constrained devices.}
    \caption{Proposed framework for fire detection using KD. The training phase involves distilling knowledge from a transformer-based teacher model (ViT/32) to the student model (MobileViT-S). The teacher model processes image patches with linear projections, positional embeddings, and a transformer encoder to produce logits, which guide the student model's learning through the distillation loss ($L_{KD}$). The student model combines convolutional layers and MobileViT modules to efficiently learn both local and global features. The trained student model is deployed on resource-constrained devices, such as drones, for real-time fire detection. The framework enables effective identification of fire regions, as illustrated by attention heatmaps generated during inference.}

    \label{fig:proposed_framework}
\end{figure*}

The challenges mentioned in the previous section must be addressed with more sophisticated approaches. We propose a framework for developing an effective, compact, and robust model for fire detection. This section describes the model architecture and the training process with KD. The overall process is depicted in Fig. \ref{fig:proposed_framework} and Algorithm \ref{alg:teacher_student_training}. The training phase begins with the pretraining of the teacher model, ViT-Base Patch32 (ViT/32), on the selected fire dataset. The teacher model learns to extract complex features through its transformer-based architecture. Then, the student model, based on MobileViT \cite{mehta2021mobilevit}, is trained using a KD framework, where the teacher model guides the student by transferring its learned knowledge. This process ensures the student model inherits the teacher's capability to recognize fire-related patterns while maintaining a more compact and lightweight architecture. Once training is complete, the trained student model is deployed on monitoring devices, such as drones. These drones patrol assigned areas, periodically capturing aerial images of the environment. The deployed model processes these images in real-time, accurately detecting fire instances. This approach enables an efficient response to potential fire hazards, which can be detected even by resource-constrained gadgets.

\subsubsection{Feature extraction and Model Architecture}

The proposed model's architecture uses transformers as convolutions \cite{mehta2021mobilevit}; in other words, by using a stack of transformers, the MobileViT module can capture global representations while also keeping the spatial order of pixels. The architecture begins with a 3×3 convolutional layer, followed by MobileNetV2 blocks, to extract local spatial features, capturing fine-grained details essential for object recognition. To model long-range dependencies and global context, the architecture utilizes MobileViT blocks. In these blocks, feature maps are unfolded into non-overlapping patches and processed using transformer layers without losing the spatial order of pixels within each patch. This approach maintains spatial inductive bias, preserving critical spatial relationships. The patches are folded back to reconstruct the feature map with local and global representations. This reconstructed feature map is projected back to a lower-dimensional space using point-wise convolutions and combined with the original features via concatenation. A final convolutional layer is then used to fuse these combined features. In fire detection, MobileViT's ability to model fine-grained details and global context enhances its capability to detect small or distant fires under challenging conditions.

\begin{algorithm}
\caption{Teacher Model Training and KD Framework}
\label{alg:teacher_student_training}
\begin{algorithmic}[1]
\REQUIRE Dataset $D = \{(x_i, y_i)\}_{i=1}^N$, Teacher model $M_t$, Student model $M_s$, Temperature $T$, Weighting factor $\alpha$, Learning rates $\eta_t$, $\eta_s$, Number of epochs $E_t$, $E_s$. 

% Step 1: Pretraining Teacher Model
\STATE \textbf{Teacher Model Training:}
\STATE Initialize $M_t$.
\FOR{epoch $e = 1$ to $E_t$}
    \STATE Shuffle dataset $D$.
    \FOR{each mini-batch $B = \{(x_b, y_b)\}$ in $D$}
        \STATE Compute teacher predictions $s_t = M_t(x_b)$.
        \STATE Calculate cross-entropy loss: $\mathcal{L}_{CE} = \frac{1}{|B|}\sum_{b=1}^{|B|} \text{CrossEntropy}(s_t, y_b)$.
        \STATE Update $M_t$ parameters using optimizer: $\theta_t \leftarrow \theta_t - \eta_t \nabla_{\theta_t} \mathcal{L}_{CE}$.
    \ENDFOR
\ENDFOR

% Step 2: KD
\STATE \textbf{KD to Student Model:}
\STATE Initialize $M_s$.
\FOR{epoch $e = 1$ to $E_s$}
    \STATE Shuffle dataset $D$.
    \FOR{each mini-batch $B = \{(x_b, y_b)\}$ in $D$}
        \STATE Compute teacher predictions $s^t = \text{Softmax}(M_t(x_b) / t)$.
        \STATE Compute student predictions $s^s = \text{Softmax}(M_s(x_b) / t)$.
        \STATE Calculate distillation loss:
        \begin{equation*}
            \mathcal{L}_{KD} = T^2 \cdot \mathcal{L}_{KLD}(s^t, s^s).
        \end{equation*}
        \STATE Compute cross-entropy loss: $\mathcal{L}_{CE} = \frac{1}{|B|}\sum_{b=1}^{|B|} \text{CrossEntropy}(s^s, y_b)$.
        \STATE Combine losses: $\mathcal{L} = (1-\alpha) \mathcal{L}_{CE} + \alpha \mathcal{L}_{KD}$.
        \STATE Update $M_s$ parameters using optimizer: $\theta_s \leftarrow \theta_s - \eta_s \nabla_{\theta_s} \mathcal{L}$.
    \ENDFOR
\ENDFOR

\RETURN $M_s$.
\end{algorithmic}
\end{algorithm}

\subsubsection{Teacher Model Architecture}

The teacher model, ViT/32, processes an input image by dividing it into non-overlapping patches, each of size 32x32 pixels, which are then linearly projected into a fixed-dimensional embedding space using a fully connected layer. To preserve the spatial order of the patches, positional encodings are added to these embeddings. The embedded patch tokens are then fed into a transformer encoder, which comprises multiple layers of multi-head self-attention and feed-forward networks. The self-attention mechanism allows the model to capture global dependencies across the entire image, enabling it to understand both local and contextual information critical for tasks like fire detection. The final output from the encoder is passed through a multilayer perceptron (MLP) head to generate logits representing the model’s predictions \cite{dosovitskiy2020image}.

\subsubsection{Knowledge Distillation} \label{section:knowledge_distillation}

KD is a technique that allows the transfer of knowledge from a complex model or an ensemble of models, known as a "teacher" model, to a simpler, smaller "student" model \cite{hinton2015distilling}. We employ KD because it is critical that the compact and fast-inference model deployed in UAVs and surveillance systems also has high performance.
In our proposed framework, we implement soft target KD as described in \cite{hinton2015distilling}.
The soft target KD involves training the student model using the teacher model's softened output probabilities (soft targets). Specifically, the total loss $\mathcal{L}$ can be expressed as:

\begin{equation}
    \mathcal{L} = (1-\alpha)\mathcal{L}_{CE}(y, y^s) + \alpha T^2 \mathcal{L}_{KLD}(s^t, s^s),
\end{equation}
where $\mathcal{L}_{CE}(y, y^s)$ is the cross-entropy loss between the true labels $y$ and the predicted probabilities of the student model $y^s$. $\mathcal{L}_{KLD}(s^t, s^s)$ is the Kullback-Leibler divergence (KLD) between the teacher's soft targets teacher $s^t$ and the student's output $s^s$, that are computed with a temperature-scaled softmax function. $T$ is the temperature parameter, and $\alpha$ is a weighting factor.

We use ViT/32 \cite{dosovitskiy2020image} as the teacher model to implement the KD techniques. This combination of the teacher model architecture and the KD technique proved the most effective based on extensive experiments. Their results can be found in subsection \ref{section:ablation_study}.

\section{Experimental Results}

This section describes the experimental setup, datasets used for evaluation, the performance and visual evaluation results, the complexity of our proposed model, and the ablation study.

\subsection{Model Implementation Details and Evaluation Metrics}

The proposed fire detection model was implemented using the PyTorch deep learning framework. We conducted the experiments on one NVIDIA A100 GPU and AMD EPYC 7402 CPU with a 2.80 GHz processor. The model was trained for 300 epochs with early stopping after 10 epochs, using a batch size of 32, and images had a resolution of 224x224. The training was done using a learning rate of 1e-4, AdamW optimizer with a weight decay of 1e-3 to prevent overfitting. 
% Augmentation techniques like Gaussian Blur were used to enhance the variability of the images and simulate visibility conditions in poor weather. 
We divided all datasets into train, validation, and test splits with 70\%, 20\%, and 10\% of images, respectively, applying the approach from previous research for fair comparison.

The performance evaluation metrics include precision (P), recall (R), F1-score (F1), and accuracy (Acc). These metrics provide a fundamental assessment of the model's effectiveness in making accurate predictions across the entire dataset \cite{yar2023effective, yar2022optimized, khan2023performance}.

\subsection{Datasets}

\begin{figure*}
    \centering

    % Row: One image from each dataset
    \begin{subfigure}{0.3\linewidth}
        \includegraphics[width=\linewidth,height=\linewidth]{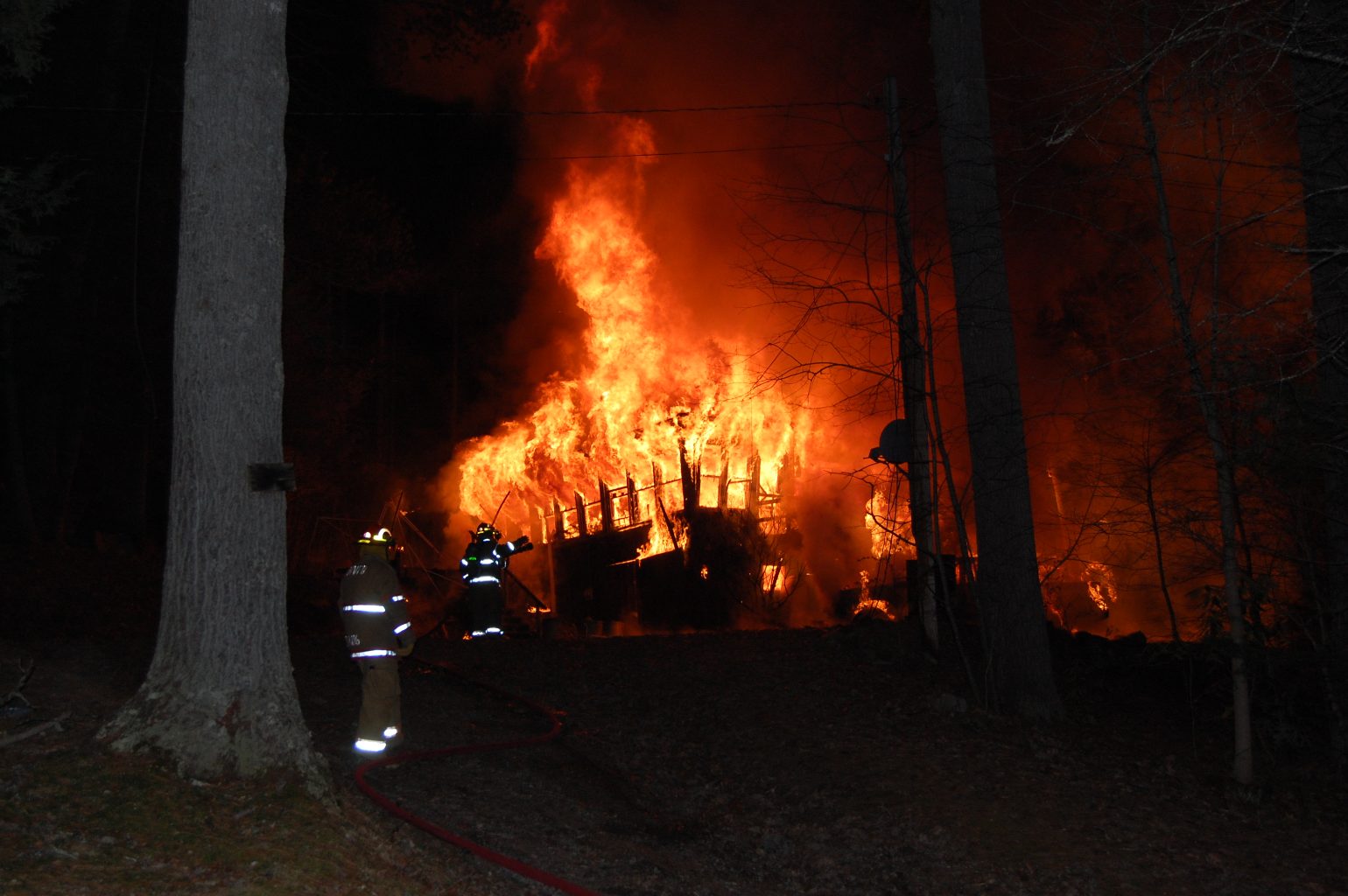}
        \caption*{BoWFire: Fire}
    \end{subfigure}
    \hfill
    \begin{subfigure}{0.3\linewidth}
        \includegraphics[width=\linewidth,height=\linewidth]{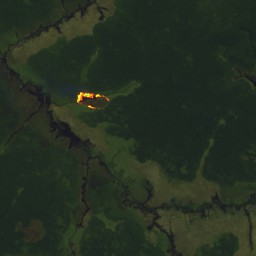}
        \caption*{ADSF: Fire}
    \end{subfigure}
    \hfill
    \begin{subfigure}{0.3\linewidth}
        \includegraphics[width=\linewidth,height=\linewidth]{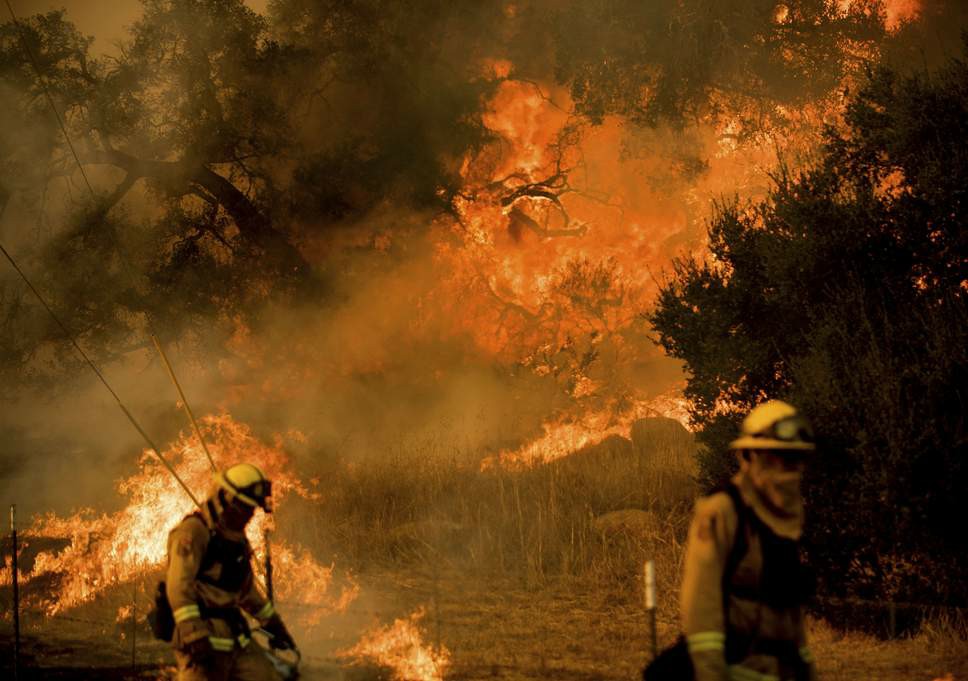}
        \caption*{DFAN: Forest Fire}
    \end{subfigure}
    
    \caption{Sample images from the fire benchmarks showcasing the diverse nature of fire detection scenarios. Each image is labeled with its respective class for training and evaluation purposes.}

    \label{fig:dataset_samples}
\end{figure*}

In this section, we present the datasets used to evaluate the performance of our model. 
Some sample images from each dataset are depicted in Fig. \ref{fig:dataset_samples}.

\subsubsection{BowFire}

The BoWFire dataset \cite{chino2015bowfire} is a small-scale fire detection dataset consisting of 119 fire images and 107 non-fire images of different resolutions. The fire images present various emergency scenarios, while non-fire images contain images without fire and images with fire-like objects (sunsets, red and yellow objects). 

% \begin{figure}[H]
%     \centering
%     \begin{subfigure}{0.47\linewidth}
%         \includegraphics[width=\linewidth, height=2.5cm]{bowfire/fire057.png}
%         \caption{Fire}
%     \end{subfigure}
%     \hfill
%     \begin{subfigure}{0.47\linewidth}
%         \includegraphics[width=\linewidth, height=2.5cm]{bowfire/fire085.png}
%         \caption{Fire}
%     \end{subfigure}
    
%     \begin{subfigure}{0.47\linewidth}
%         \includegraphics[width=\linewidth, height=2.5cm]{bowfire/not_fire007.png}
%         \caption{Not fire}
%     \end{subfigure}
%     \hfill
%     \begin{subfigure}{0.47\linewidth}
%         \includegraphics[width=\linewidth, height=2.5cm]{bowfire/not_fire011.png}
%         \caption{Not fire}
%     \end{subfigure}
%     \caption{BoWFire dataset: example images from different categories.}
%     \label{fig:fire_images}
% \end{figure}

\subsubsection{ADSF}

% Most existing datasets focused on images from CCTV cameras and synthetic 3D images. However, the surveillance systems are diverse and use many devices. 
The ASDF dataset was introduced in \cite{yar2023effective}, containing images from drones and satellites. This dataset consists of 3000 fire images and 3000 normal images shot outdoors. The ADSF dataset provides a range of images in different conditions, such as time of the day, landscape, and altitude. 

% \begin{figure}[H]
%     \centering
%     \begin{subfigure}{0.47\linewidth}
%         \includegraphics[width=\linewidth, height=2.5cm]{adsf/Fire (2494).jpg}
%         \caption{Fire}
%     \end{subfigure}
%     \hfill
%     \begin{subfigure}{0.47\linewidth}
%         \includegraphics[width=\linewidth, height=2.5cm]{adsf/Fire (9).jpg}
%         \caption{Fire}
%     \end{subfigure}
    
%     \begin{subfigure}{0.47\linewidth}
%         \includegraphics[width=\linewidth, height=2.5cm]{adsf/Normal.jpg}
%         \caption{Normal}
%     \end{subfigure}
%     \hfill
%     \begin{subfigure}{0.47\linewidth}
%         \includegraphics[width=\linewidth, height=2.5cm]{adsf/Normal (7).jpg}
%         \caption{Normal}
%     \end{subfigure}
%     \caption{ADSF dataset: example images from different categories.}
%     \label{fig:fire_images}
% \end{figure}

\subsubsection{DFAN}
The DFAN dataset is a medium-scale dataset that consists of 3,804 images of different fire scenarios, split into 12 imbalanced classes. Proposed by Yar \textit{et al.} \cite{yar2022optimized}, this dataset challenges models with the diversity of classes. Training a model on this dataset allows us to identify the characteristics of the fire and respond to it according to the level of the crisis.

\subsection{Visual Results}
\begin{figure}[h]
    \centering
    \includegraphics[width=\linewidth]{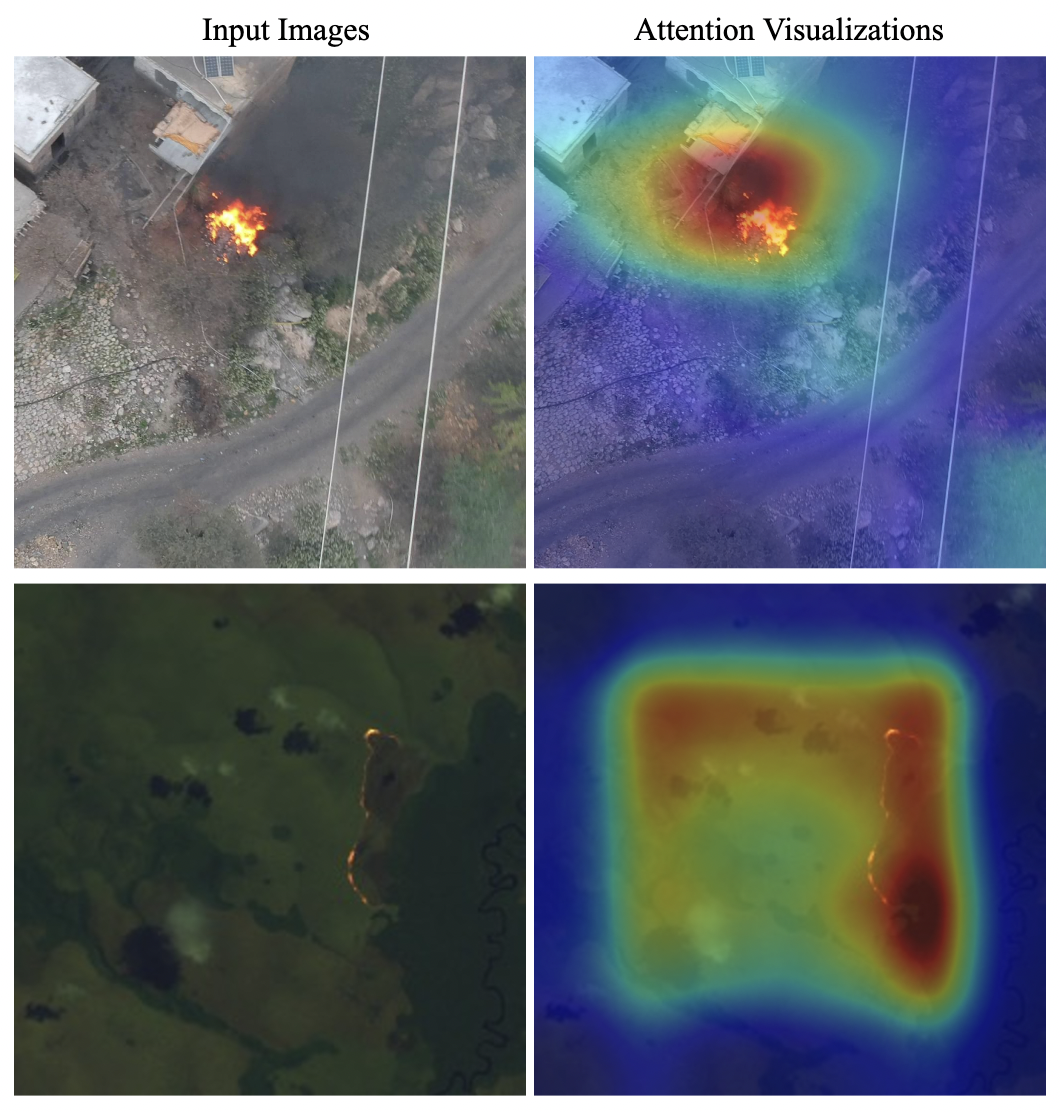}
    % \caption{Visualization of the model attention on the drone and satellite images.}
    \caption{Visualization of the model attention on drone and satellite images. The left column displays the input images, while the right column presents the corresponding Grad-CAM-based attention visualizations. The top row shows a fire in an urban environment captured by a drone, with the attention map clearly highlighting the fire region amidst surrounding objects. The model is able to effectively focus on fire regions across diverse environmental conditions and input modalities.}

    \label{fig:adsf_visualization}
\end{figure}

The visual results showcased in Fig. \ref{fig:adsf_visualization} demonstrate the effectiveness of the proposed model in localizing fire regions across diverse environments. The left column displays the input images, while the right column presents the corresponding attention heatmaps visualized with Grad-CAM. Input images and attention heatmaps demonstrate the model's ability to detect flames in drone and satellite imagery, even in challenging scenarios. For instance, in the first row, the model correctly highlights the area of active flames in a satellite image of a building fire. Similarly, in the second row, the model successfully identifies fire spread over vegetation in a drone-captured image. However, limitations are observed, such as misinterpreting clouds as fire smoke due to their visual similarity. This signifies the need for further improvement in distinguishing fire-related features from non-fire elements in complex scenes.

\begin{figure}[h]
    \centering
    % Row 1: Images 1, 2, 3
    \begin{subfigure}{0.3\linewidth}
        \centering
        \includegraphics[width=\linewidth]{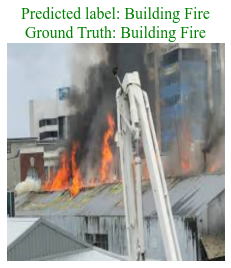}
        \label{fig:image1}
    \end{subfigure}
    \hfill
    \begin{subfigure}{0.3\linewidth}
        \centering
        \includegraphics[width=\linewidth]{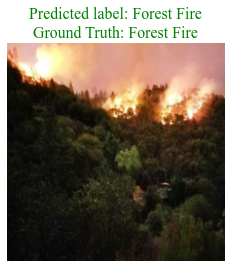}
        \label{fig:image2}
    \end{subfigure}
    \hfill
    \begin{subfigure}{0.3\linewidth}
        \centering
        \includegraphics[width=\linewidth]{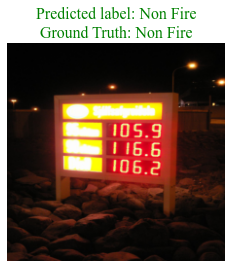}
        \label{fig:image3}
    \end{subfigure}
    % \vspace{-0.5cm}
    \begin{subfigure}{0.3\linewidth}
        \centering
        \includegraphics[width=\linewidth]{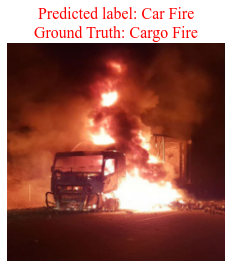}
        \label{fig:image4}
    \end{subfigure}
    \hfill
    \begin{subfigure}{0.3\linewidth}
        \centering
        \includegraphics[width=\linewidth]{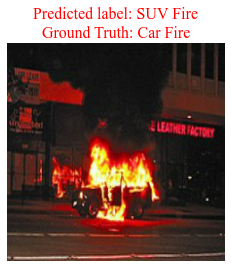}
        \label{fig:image5}
    \end{subfigure}
    \hfill
    \begin{subfigure}{0.3\linewidth}
        \centering
        \includegraphics[width=\linewidth]{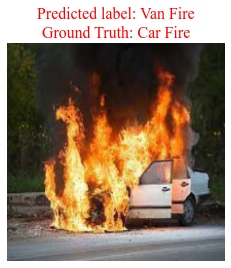}
        \label{fig:image6}
    \end{subfigure} 
    \vspace{-0.5cm}
    % \caption{Demonstration of the correctly and incorrectly labeled DFAN images.}
    \caption{Demonstration of correctly and incorrectly labeled DFAN images. The top row displays correctly classified examples, including a "Building Fire," "Forest Fire," and "Non-Fire" scene. The bottom row presents misclassified examples, where a "Cargo Fire" was predicted as "Car Fire," an "SUV Fire" was correctly labeled as "Car Fire," and a "Van Fire" was predicted as "Car Fire." }

    \label{fig:correctly_incorrectly_labeled}
\end{figure}

Moreover, Fig. \ref{fig:correctly_incorrectly_labeled} displays sample images from the DFAN dataset, the predictions made by our proposed model, and the ground truth labels for each image. Our model clearly differentiates between visually unalike classes but can make mistakes in related classes. For example, "Car Fire", "SUV Fire", and "Van Fire" classes are often confused with each other. These examples highlight the model's challenges in handling visually related categories, particularly in scenarios where subtle differences in object shape or fire intensity can mislead predictions.

The visual evaluation provides valuable insights into the possible improvement directions of the model.

\subsection{Ablation study} \label{section:ablation_study}

We conducted numerous experiments to distill knowledge from stronger models to improve our proposed model's performance on ADSF and DFAN datasets. The knowledge techniques used in the experiments include soft target KD \cite{hinton2015distilling}, Distillation from A Stronger Teacher (DIST) \cite{huang2022knowledge}, and One-for-All KD (OFA-KD) \cite{hao2024one}. 
% The principle behind the soft target knowledge distillation was described in subsection \ref{section:knowledge_distillation}, and we explored two other knowledge distillation techniques due to their reported positive improvement, especially in cases with stronger teachers and heterogeneous models. DIST preserves relational information between the teacher and student outputs, making it better suited for training students under the guidance of stronger teacher models. The student model does not overfit the teacher's outputs but strives to preserve inter-class and intra-class relations. OFA-KD is a knowledge distillation technique for cases where the teacher and student models belong to different model families: convolutional neural networks (CNNs), transformers, or multi-layer perceptrons (MLPs). In OFA-KD, intermediate features are projected into a shared latent space, such as the logits space, where architecture-specific information is discarded. This allows for effective distillation even when the teacher and student models have different inductive biases \cite{hao2024one}. 

Moreover, we employed ViT/32 \cite{dosovitskiy2020image} from the transformers family and ConvNeXt-Base (ConvNeXt) \cite{liu2022convnet} from the CNN family as the teacher models. These architectures were selected for their ability to provide different knowledge to our proposed model. The student models tested included MobileViT-S and MobileViT-XS to examine the effects of model size on performance. The baseline performance of MobileViT-S on the test splits of the DFAN and ADSF datasets without KD is 90.29\% and 95.50\%, respectively. For MobileViT-XS, the performance is 87.93\% and 94.00\%. 

Given resource constraints, we avoided an exhaustive grid search for hyperparameter optimization. Instead, we incrementally optimized individual parameters based on their observed impact on performance. For soft target KD, we found that $T=2$ and $\alpha=0.1$ provided the best balance between the distillation loss and the standard cross-entropy loss. For DIST, we used $\alpha=0.1$, $\beta=2$, $\gamma=2$, and $\tau=1$, while for OFA-KD, the optimal parameters were $\epsilon=1.2$ and $T=3$, as described in their corresponding papers \cite{huang2022knowledge, hao2024one}. 
% These parameters were selected through iterative refinement and were not further adjusted due to the computational cost of a comprehensive search.

\begin{table*}[htb]
\centering
% \caption{Performance of our model with different knowledge distillation techniques on DFAN and ADSF datasets.}
\caption{Performance comparison of our model using different KD techniques on the fire benchmarks. The table highlights the accuracies achieved by the MobileViT-S and MobileViT-XS student models under three distillation techniques: Soft Target KD, DIST, and OFA. The results in bold signify the best accuracies for each model architecture.}

\label{tab:merged_kd_results}
\resizebox{\textwidth}{!}{%
\begin{tabular}{l|l|cc|cc|cc}
\hline
\textbf{Dataset} & \textbf{Teacher} & \multicolumn{2}{c|}{\textbf{Soft Target KD}} & \multicolumn{2}{c|}{\textbf{DIST}} & \multicolumn{2}{c}{\textbf{OFA}} \\
% \hline
 & & MobileViT-S & MobileViT-XS & MobileViT-S & MobileViT-XS & MobileViT-S & MobileViT-XS \\
\hline
\multirow{2}{*}{DFAN} 
& ViT/32 & \textbf{91.08} & 90.55 & 89.50 & 89.24 & 89.76 & 86.09 \\
& ConvNeXt & 88.98 & 88.71 & 90.55 & 89.76 & 88.71 & 88.45 \\
\hline
\multirow{2}{*}{ADSF} 
& ViT/32 & 91.33 & 94.83 & 94.17 & 93.00 & 95.33 & 95.00 \\
& ConvNeXt & 92.00 & 92.33 & 95.00 & 95.00 & 95.17 & \textbf{95.50} \\
\hline
\end{tabular}
}
\end{table*}

Table \ref{tab:merged_kd_results} highlights that KD significantly enhances the performance of MobileViT-S on the DFAN dataset. The best result, an accuracy of 91.08\%, was achieved using soft target KD with ViT/32 as the teacher model. This improvement underscores the importance of global contextual knowledge provided by the transformer-based teacher. In comparison, MobileViT-XS achieves slightly lower performance, with a maximum accuracy of 90.55\% under the same configuration. This demonstrates that while MobileViT-XS is lightweight, it is less effective in handling the complex scenarios present in the DFAN dataset. OFA-KD and DIST also improved performance compared to the baseline but showed slightly lower results than soft target KD, likely due to the specific properties of the DFAN dataset, which benefits from the global context provided by ViT/32. The results also reveal that ConvNeXt, a CNN-based teacher, does not provide as much performance improvement as ViT/32. For instance, the best accuracy achieved with ConvNeXt as the teacher was 88.98\% for MobileViT-S, indicating that the global feature extraction of ViT/32 is better suited for challenging datasets like DFAN.

On the ADSF dataset, MobileViT-XS achieves the best accuracy of 95.50\% using OFA-KD with ConvNeXt as the teacher. This result slightly surpasses MobileViT-S, which achieves a maximum accuracy of 95.33\% under the same configuration. The ADSF dataset, with only two classes, is relatively simpler than DFAN, making it less reliant on the global contextual features provided by ViT/32. Consequently, the lightweight MobileViT-XS model performs competitively on this dataset. Interestingly, the baseline accuracy for MobileViT-S on ADSF is already 95.50\%, indicating that the dataset’s simplicity limits the impact of KD. 

While MobileViT-XS achieves competitive performance on the ADSF dataset, MobileViT-S outperforms it on the more challenging DFAN dataset, with an accuracy of 91.08\% compared to 90.55\%. This suggests that MobileViT-S is better suited for complex scenarios requiring robust feature extraction and generalization. 
% Furthermore, the marginal improvements of MobileViT-XS on ADSF come at the cost of reduced performance on DFAN, highlighting its limitations in handling datasets with fine-grained class distinctions.

% Given these findings, we selected MobileViT-S as the backbone for our proposed framework due to its consistent performance across datasets and its superior handling of complex fire detection scenarios. Additionally, as discussed in subsection \ref{section:complexity_analysis}, MobileViT-S already exhibits SOTA inference speed on all tested devices, making it a practical choice for deployment.

\subsection{Performance Evaluation}

In this section, we compare the performance and the complexity of our proposed model with the existing solutions. The methods are compared on the datasets described above.

\subsubsection{Performance on the Evaluation Metrics}

Table \ref{tab:combined_comparison} compares the performance of our proposed model to existing methods on the three fire datasets. The evaluation highlights the effectiveness of our approach across multiple scenarios, showcasing both strengths and areas for improvement. On the BoWFire dataset, our model achieves perfect scores across all metrics even without implementing KD, with 100\% Acc, F1, Rec, and Pre, demonstrating its exceptional capability to generalize on this dataset. In comparison, previous SOTA methods, such as MAFire-Net \cite{khan2023enhancing}, achieved strong results with 97.82\% Acc and an F1 of 97.77\%, but our model still outperforms them. However, it is important to note that the small size of the BoWFire dataset limits its representativeness and may lead to inflated performance metrics. Due to this, we further evaluated our model on other datasets.

% \begin{table}
% \centering
% \caption{Comparison of the performance of our proposed model to the existing works on the BoWFire dataset.}
% \label{tab:bowfire_comparison}
% \begin{tabular}{lcccc}
% \hline
% \textbf{Methods} & \textbf{Acc} & \textbf{F1} & \textbf{Rec} & \textbf{Pre} \\
% \hline
% EFDNet \cite{li2020efficient} & 83.33 & 81.85 & 83.00 & 81.81 \\
% ANetFire \cite{muhammad2018early} & 88.05 & 88.00 & 98.00 & 80.00 \\
% Xception \cite{sathishkumar2023forest} & 91.41 & - & - & - \\
% EMNFire \cite{muhammad2019efficient} & 92.04 & 92.00 & 93.00 & 90.00 \\
% DFAN (comp.) \cite{yar2022optimized} & 93.00 & 93.10 & 92.00 & 94.30 \\
% DFAN \cite{yar2022optimized} & 95.00 & 95.00 & 94.00 & 95.00 \\
% OFAN \cite{dilshad2024toward} &  96.23 & 96.00 & 95.00 & 96.00 \\
% MAFire-Net \cite{khan2023enhancing} & 97.82 & 97.77 & 98.15 & 97.05 \\
% FireClassNet \cite{daoud2023fireclassnet} & 99.56 & 99.58 & 99.44 & 99.72 \\
% Our Model & 100 & 100 & 100 & 100 \\
% \hline
% \end{tabular}
% \end{table}

\begin{figure*}
    \centering
    \begin{subfigure}[b]{0.45\linewidth}
        \centering
        \includegraphics[height=6cm]{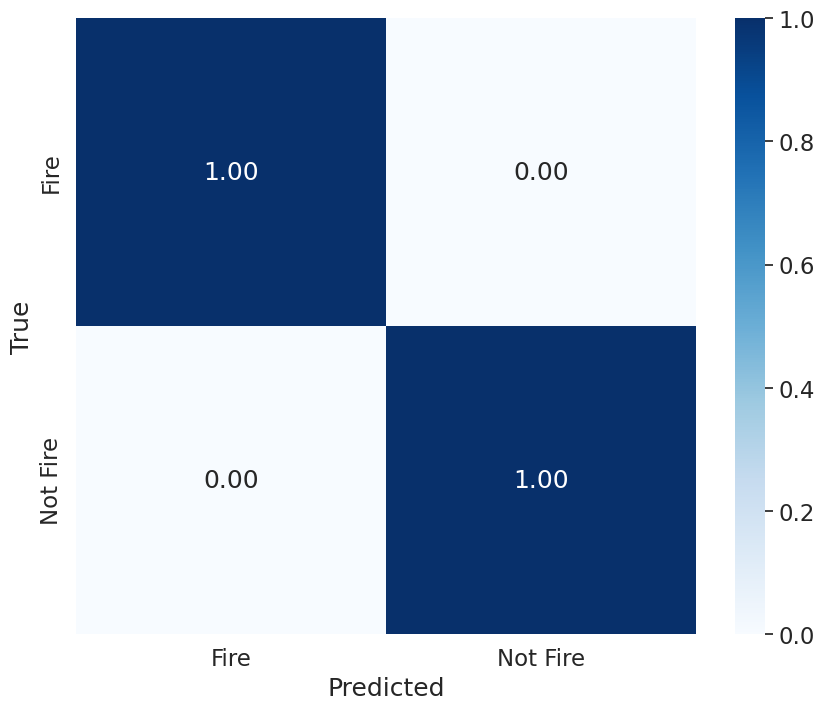}
        \caption{BoWFire dataset.}
        \label{fig:bowfire_confusion_matrix}
    \end{subfigure}
    \hfill
    \begin{subfigure}[b]{0.45\linewidth}
        \centering
        \includegraphics[height=6cm]{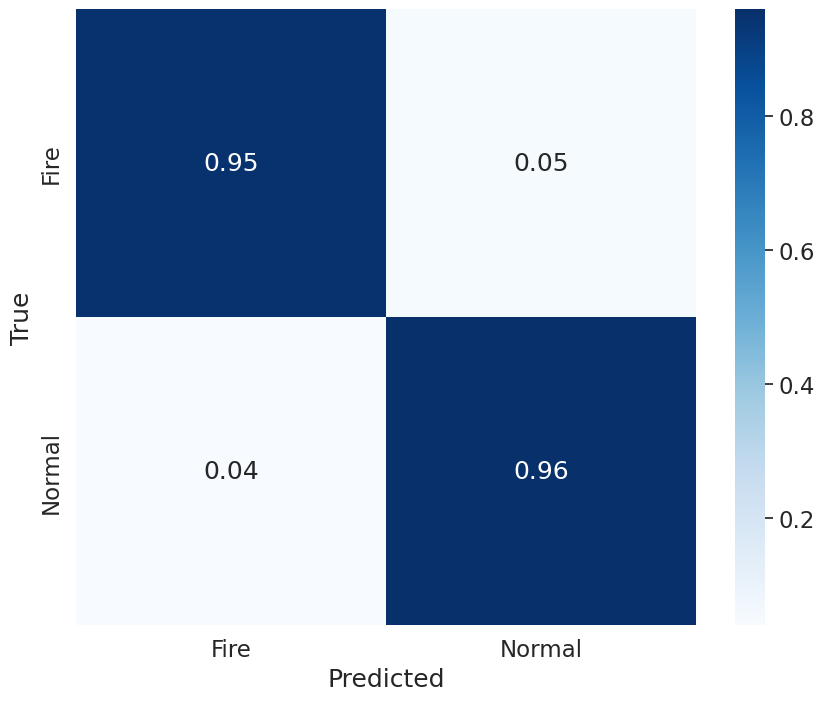}
        \caption{ADSF dataset.}
        \label{fig:adsf_confusion_matrix}
    \end{subfigure}
    
    \vskip 0.5cm % Space between rows
    
    \begin{subfigure}[b]{0.8\linewidth}
        \centering
        \includegraphics[height=6cm]{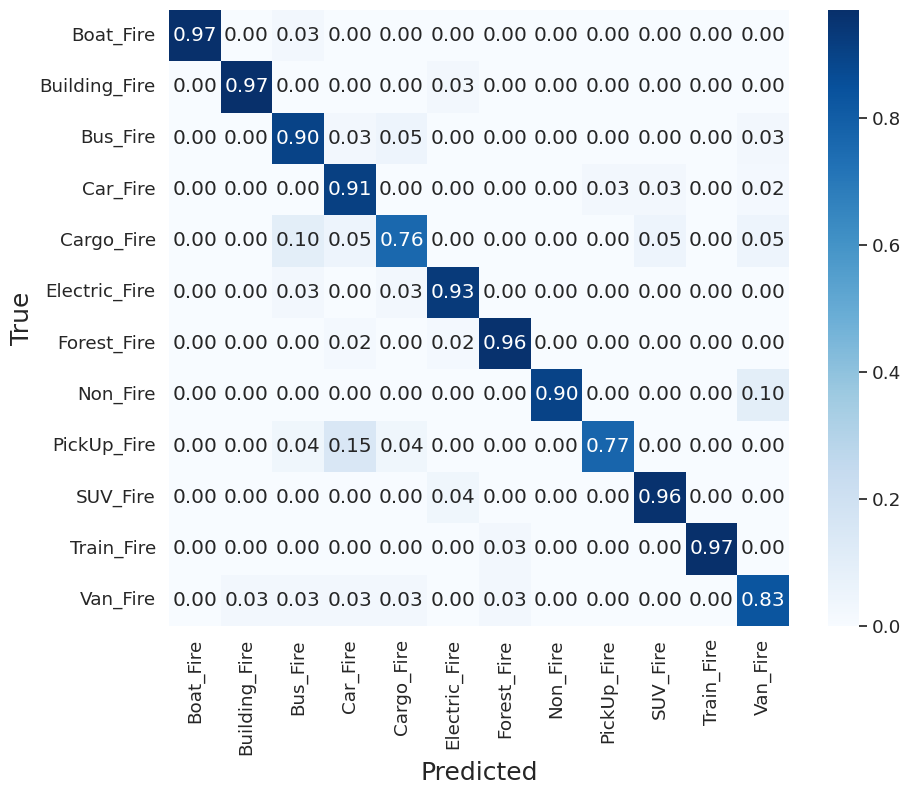}
        \caption{DFAN dataset.}
        \label{fig:dfan_confusion_matrix}
    \end{subfigure}

    \caption{Confusion matrices for the fire benchmarks. For the BoWFire dataset, the model achieves perfect classification with no misclassifications. On the ADSF dataset, the confusion matrix demonstrates high accuracy, with minor misclassifications between fire and non-fire classes. The DFAN dataset's confusion matrix captures the complexity of multiclass fire detection, with most classes achieving high classification accuracy, but for some classes, the accuracy falls behind, such as "Car Fire," "SUV Fire," and "Van Fire".}

    \label{fig:merged_confusion_matrices}
\end{figure*}

Fig. \ref{fig:adsf_confusion_matrix} shows the confusion matrix of our proposed model on the test split of the ADSF dataset. The results demonstrate that our model outperforms all previous works across all metrics, achieving an Acc, F1, Rec, and Pre of 95.50\%. Among the existing methods, MobileNetV3 + MSAM \cite{yar2024efficient} shows strong performance, with an accuracy and F1-score of 93.50\% and 93.51\%, respectively. However, our model surpasses this by a margin of 2.0\%, reflecting its stronger capability in fire detection tasks. Overall, the consistent superiority of our model across all metrics demonstrates its robustness and effectiveness in accurately identifying fire regions under diverse scenarios of the ADSF dataset.

\begin{table*}[h]
\centering
\caption{Comparison of the performance of our proposed model to the existing work across the fire benchmarks. The metrics in bold represent the best performance, while the underscored metrics are the second-best performance.}
\label{tab:combined_comparison}
\resizebox{\textwidth}{!}{%
\begin{tabular}{l|cccc|cccc|cccc}
\hline
\multirow{2}{*}{\textbf{Methods}} & \multicolumn{4}{c|}{\textbf{BoWFire Dataset}} & \multicolumn{4}{c|}{\textbf{ADSF Dataset}} & \multicolumn{4}{c}{\textbf{DFAN Dataset}} \\
 & \textbf{Acc} & \textbf{F1} & \textbf{Rec} & \textbf{Pre} & \textbf{Acc} & \textbf{F1} & \textbf{Rec} & \textbf{Pre} & \textbf{Acc} & \textbf{F1} & \textbf{Rec} & \textbf{Pre} \\
\hline
EFDNet \cite{li2020efficient} & 83.33 & 81.85 & 83.00 & 81.81 & 88.00 & 87.75 & 88.00 & 87.50 & 77.50 & 77.49 & 77.00 & 78.00 \\
ANetFire \cite{muhammad2018early} & 88.05 & 88.00 & 98.00 & 80.00 & - & - & - & - & - & - & - & - \\
Xception \cite{sathishkumar2023forest} & 91.41 & - & - & - & - & - & - & - & - & - & - & - \\
EMNFire \cite{muhammad2019efficient} & 92.04 & 92.00 & 93.00 & 90.00 & - & - & - & - & - & - & - & - \\
DFAN (comp.) \cite{yar2022optimized} & 93.00 & 93.10 & 92.00 & 94.30 & - & - & - & - & 86.50 & 86.00 & 87.00 & 86.00 \\
DFAN \cite{yar2022optimized} & 95.00 & 95.00 & 94.00 & 95.00 & 89.36 & 89.84 & 94.00 & 86.01 & 88.00 & 87.00 & 88.00 & 88.00 \\
OFAN \cite{dilshad2024toward} & 96.23 & 96.00 & 95.00 & 96.00 & - & - & - & - & - & - & - & - \\
MAFire-Net \cite{khan2023enhancing} & 97.82 & 97.77 & 98.15 & 97.05 & - & - & - & - & 88.83 & 87.53 & 86.44 & 89.35 \\
FireClassNet \cite{daoud2023fireclassnet} & \underline{99.56} & \underline{99.58} & \underline{99.44} & \underline{99.72} & - & - & - & - & - & - & - & - \\
ResNet50 + FAN \cite{yar2022optimized} & - & - & - & - & - & - & - & - & 86.12 & 85.00 & 86.00 & 88.00 \\
NASNetM + FAN \cite{yar2022optimized} & - & - & - & - & - & - & - & - & 82.56 & 81.00 & 82.00 & 82.00 \\
MobileNet + FAN \cite{yar2022optimized} & - & - & - & - & - & - & - & - & 85.30 & 85.00 & 85.00 & 85.00 \\
ADFireNet \cite{yar2023effective} & - & - & - & - & 90.86 & 89.84 & 90.86 & 90.90 & 90.00 & 89.99 & \underline{90.49} & \underline{90.43} \\
MobileNetV3 + MSAM \cite{yar2024efficient} & - & - & - & - & \underline{93.50} & \underline{93.51} & \underline{93.51} & \underline{93.57} & \textbf{91.20} & \underline{90.63} & \textbf{91.17} & 90.36 \\
Our Model & \textbf{100} & \textbf{100} & \textbf{100} & \textbf{100} & \textbf{95.50} & \textbf{95.50} & \textbf{95.50} & \textbf{95.50} & \underline{91.08} & \textbf{90.75} & 90.27 & \textbf{91.43} \\
\hline
\end{tabular}
}
\end{table*}

As seen from Table \ref{tab:combined_comparison}, our model achieves an accuracy of 91.08\%, an F1-score of 90.75\%, a recall of 90.27\%, and a precision of 91.43\%. While these results position our model competitively among existing works, it falls slightly short in some metrics. Specifically, our model achieves the second-best result in accuracy, with a 0.12\% gap between MobileNetV3 + MSAM \cite{yar2024efficient}. Our model shows the best F1 and precision scores, but ADFireNet \cite{yar2023effective} and MobileNetV3 + MSAM have higher recall of 90.49\% and 91.17\%, respectively. Notably, our model outperforms earlier approaches, indicating improved generalization compared to earlier architectures. Insights from the confusion matrix in Fig. \ref{fig:dfan_confusion_matrix} further validate the robustness of our model. The matrix highlights its ability to accurately classify critical fire categories like "Forest Fire" and "Car Fire," achieving high classification counts in these categories. However, limitations are observed in classes like "Cargo Fire" and "Pickup Fire," where some misclassifications occur, potentially due to visual similarities with other categories. This underscores an area for improvement in further refining the model's attention mechanism to reduce misclassification of visually similar categories.

% \begin{figure}[H]
%     \centering
%     \includegraphics[width=0.7\linewidth]{confusion_matrix/dfan_confusion_matrix.png}
%     \caption{Confusion matrix on the ADSF dataset}
%     \label{fig:dfan_confusion_matrix}
% \end{figure}

\subsubsection{Complexity Analysis} \label{section:complexity_analysis}

Due to the resource constraints of surveillance systems, UAVs, and IoT devices, the fire detection model should be able to quickly predict the class of an image on any system. Table \ref{tab:complexity_analysis} presents the system specifications, model size, and frames-per-second (FPS) of our proposed model and the existing work.

The proposed model demonstrates significant improvements in performance compared to existing methods, as shown in Table \ref{tab:complexity_analysis}. It achieved an impressive 431.07 FPS on an Nvidia A100, 28.04 FPS on an AMD EPYC 7402 (2.80 GHz), and 9.36 on Raspberry Pi 4 with a compact model size of 19.73 MB. In comparison, EMNFire \cite{muhammad2019efficient} has the smallest model size of 13.0 MB, but its FPS values were lower, achieving 34.0 on a TITAN X (12GB) and 5.0 on a Raspberry Pi. While MobileNetV3 + MSAM \cite{yar2024efficient} provides competitive FPS on high-performance systems (75.15 FPS on a GeForce RTX-3090), it falls behind on Raspberry Pi with 8.0 FPS and has a larger model size of 25.20 MB. Similarly, DFAN (compressed) \cite{yar2022optimized} achieves a high 125.33 FPS on an RTX 2070, but its larger model size of 41.09 MB makes it less suitable for resource-constrained devices, where our model delivers a better balance of compactness and speed.
Despite the strong performance of these models, the proposed model outperformed all others in terms of FPS and model size. It achieved the highest FPS values on all system specifications while maintaining a smaller model size, demonstrating its efficiency and scalability for deployment on various devices, including resource-constrained environments.

\begin{table}[h]
\centering
\caption{Complexity analysis of the proposed model compared with existing research on different devices.}
\label{tab:complexity_analysis}
\resizebox{\linewidth}{!}{%
\begin{tabular}{l|l|c|c}
\hline
\textbf{Model} & \textbf{System} & \textbf{Size (MB)} & \textbf{FPS} \\
\hline
\multirow{2}{*}{EMNFire \cite{muhammad2019efficient}} & TITAN X (12GB) & \multirow{2}{*}{13.0} & 34.0 \\ 
 & Raspberry Pi & & 5.0 \\ 
\hline
\multirow{2}{*}{GNetFire \cite{muhammad2018convolutional}} & TITAN X (12GB) & \multirow{2}{*}{43.3} & 20.0 \\ 
 & Raspberry Pi & & 4.0 \\ 
\hline
\multirow{2}{*}{SE-EFFNet \cite{khan2022randomly}} & RTX 2070 (12GB) & \multirow{2}{*}{47.75} & 45.0 \\ 
 & Raspberry Pi & & 6.0 \\ 
\hline
\multirow{3}{*}{DFAN (comp.) \cite{yar2022optimized}} & RTX 2070 (12GB) & \multirow{3}{*}{41.09} & 125.33 \\ 
 & Intel i9 (3.60GHz) & & 22.73 \\ 
 & Raspberry Pi & & 3.21 \\ 
\hline
\multirow{2}{*}{OFAN \cite{dilshad2024toward}} & Intel i9 (5.00GHz) & \multirow{2}{*}{12.20} & 25.50 \\ 
 & Raspberry Pi & & 8.37 \\ 
\hline
\multirow{3}{*}{MAFire-Net \cite{khan2023enhancing}} & GeForce RTX-3090 & \multirow{3}{*}{74.43} & 78.31 \\ 
 & Intel i10 (5.3GHz) & & 14.32 \\ 
 & Raspberry Pi & & 0.92 \\ 
\hline
\multirow{3}{*}{\begin{tabular}[c]{@{}c@{}}MobileNetV3 \\ + MSAM \cite{yar2024efficient}\end{tabular}} & GeForce RTX-3090 & \multirow{3}{*}{25.20} & 75.15 \\ 
 & Intel i9 (3.60GHz) & & 24.0 \\ 
 & Raspberry Pi & & 8.0 \\ 
\hline
\multirow{3}{*}{Our Model} & A100 & \multirow{3}{*}{19.73} & 431.07 \\ 
 & EPYC 7402 (2.80 GHz) & & 28.04 \\ 
 & Raspberry Pi & & 9.36 \\ 
\hline
\end{tabular}
}
\end{table}

\section{Conclusion}
In this work, we proposed a lightweight and efficient fire detection model based on the MobileViT-S architecture, optimized through KD techniques to achieve high accuracy and real-time inference on resource-constrained devices. By leveraging the inherent hybrid structure of MobileViT-S, which combines the local feature extraction capabilities of CNNs with the global context modeling of transformers, our model demonstrates exceptional performance in detecting fire and wildfire regions under diverse surveillance conditions. Through rigorous experiments on benchmark datasets such as BoWFire, ADSF, and DFAN, the proposed model achieved 100\%, 95.50\%, and 91.08\% accuracies and lowered false positive rate. Notably, the model not only surpassed or matched SOTA results but also achieved the highest FPS across all tested devices, demonstrating its suitability for real-time applications.

Nevertheless, our approach has some limitations. First, the model's ability to differentiate between visually similar elements, such as smoke and clouds, needs improvement to minimize false positives. Second, exploring advanced data augmentation techniques or incorporating temporal information from video sequences could enhance the model's generalization capability in dynamic environments. Lastly, future work could investigate more sophisticated KD strategies to better utilize diverse teacher models and further improve the student's performance. By addressing the identified limitations and exploring the proposed directions, the robustness and feasibility of fire monitoring systems can be further enhanced.

\bibliographystyle{IEEEtran}
\bibliography{references}

\vfill

\end{document}